%% file: root.tex
\documentclass[letterpaper, 10 pt, conference]{ieeeconf}  

\IEEEoverridecommandlockouts                              

\usepackage{subcaption}
\usepackage{graphicx}
\usepackage{hyperref}
\usepackage{tabularx}
\usepackage{multirow}
\let\labelindent\relax
\usepackage{enumitem}
\usepackage{balance}

\title{\LARGE \bf
How Much is too Much: Exploring the Effect
of \\ Verbal Route Description Length on Indoor Navigation
}

\author{Fathima Nourin N$^{1}$, Pradip Pramanick$^{1, 2}$, and Chayan Sarkar$^{1}$
\thanks{$^{1}$Fathima Nourin N \& Chayan Sarkar are with TCS Research, India
        {\tt\small \{fathimanourin.n, sarkar.chayan\}@tcs.com}}%
\thanks{$^{2}$Pradip Pramanick is with Interdepartmental Center for Advances in Robotic Surgery - ICAROS, University of Naples Federico II, Naples, Italy {\tt\small pradip.pramanick@unina.it}}%
}

\begin{document}

\maketitle
\thispagestyle{empty}
\pagestyle{empty}

\begin{abstract}
Navigating through a new indoor environment can be stressful. Recently, many places have deployed robots to assist visitors. One of the features of such robots is escorting the visitors to their desired destination within the environment, but this is neither scalable nor necessary for every visitor. Instead, a robot assistant could be deployed at a strategic location to provide wayfinding instructions. This not only increases the user experience but can be helpful in many time-critical scenarios e.g., escorting someone to their boarding gate at an airport. However, delivering route descriptions verbally poses a challenge. If the description is too verbose, people may struggle to recall all the information, while overly brief descriptions may be simply unhelpful. This article focuses on studying the optimal length of verbal route descriptions that are effective for reaching the destination and easy for people to recall. This work proposes a theoretical framework that links route segments to chunks in working memory. Based on this framework, an experiment is designed and conducted to examine the effects of route descriptions of different lengths on navigational performance. The results revealed intriguing patterns suggesting an ideal length of four route segments. This study lays a foundation for future research exploring the relationship between route description lengths, working memory capacity, and navigational performance in indoor environments. 

\end{abstract}

\input{01_intro}

\input{02_methodology}
\input{03_results}
\input{05_future_work}
\input{04_conclusion}


\balance
\bibliographystyle{IEEEtran}
\bibliography{root}

\end{document}

%% file: 01_intro.tex
\section{Introduction}
Over the past few years, the study of navigation and wayfinding in indoor environments has emerged as an important research area~\cite{1, 2}. It is driven by the growing recognition of its profound implications for human mobility, accessibility, and spatial cognition. Traditional geospatial science and popular wayfinding \& navigation systems focus on outdoor environments. But humans spent most (about 90\%) of their time in indoor environments~\footnote[3]{https://essay.utwente.nl/85763/}. Large indoor environments, like hospitals, schools, offices, shopping malls, airports, railway stations, etc., constitute the primary spaces where most humans spend a significant portion of their time throughout their lives. Thus, indoor wayfinding is a common scenario for many people in their daily lives. A study by Viaene~\cite{viaene2018} notes that during the experiments, participants consistently reported facing more difficulties with indoor wayfinding compared to outdoor navigation. Moreover, indoor environments present unique challenges due to their architectural structures like elevation changes, limited field of view \& poor visual access. These factors make it difficult to find and focus on key objects for orientation~\cite{viaene2018}. Thus, the indoor wayfinding experience is complex and markedly different from the outdoor wayfinding experience. 

\begin{figure}
    \centering
    \includegraphics[width=\linewidth]{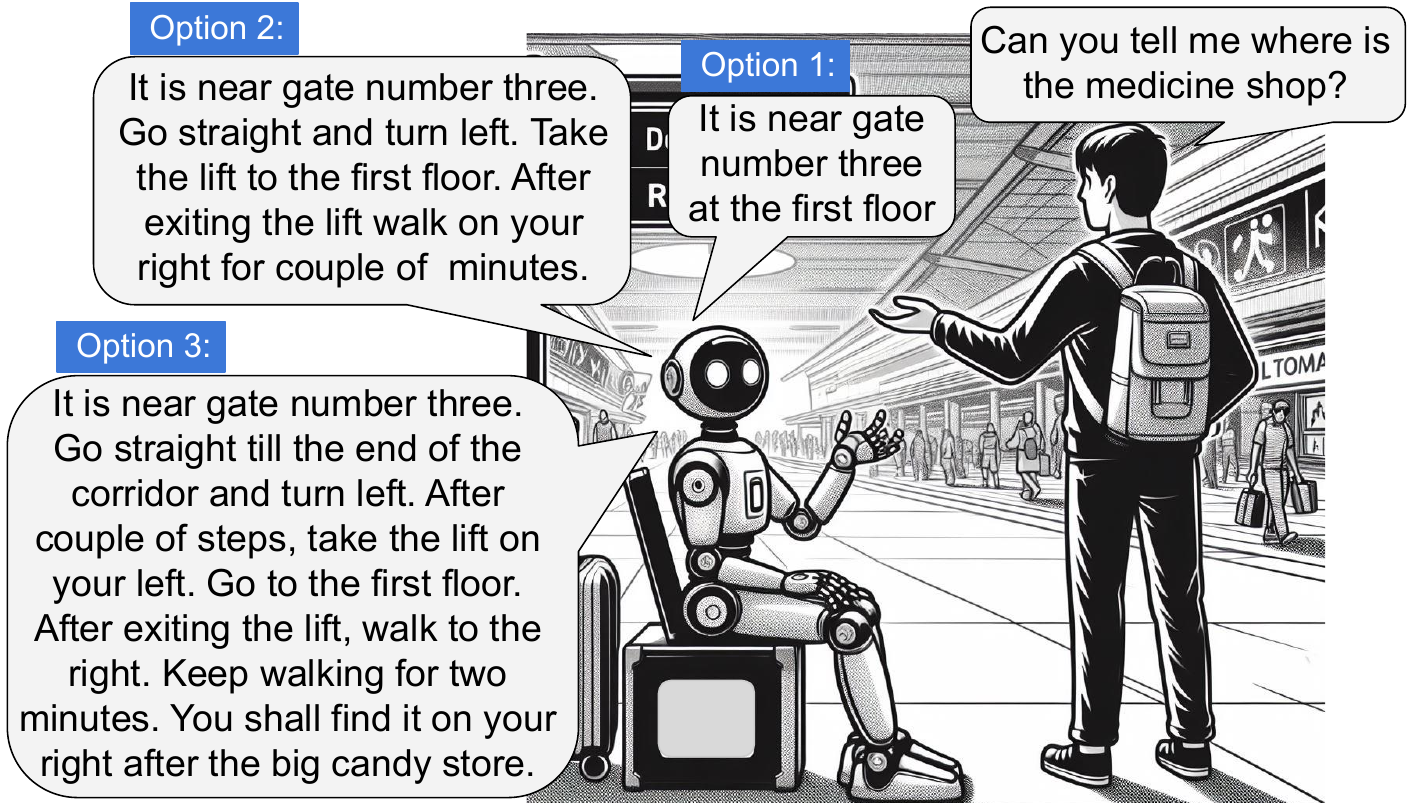}
.    \caption{Different lengths of verbal route descriptions for the same route -- a very crisp description may not help find the destination (option 1), whereas a very detailed description may be difficult to remember (option 3).}
    \label{fig:route_direction_example}
    \vspace{-1em}
\end{figure}

Navigating through complex indoor spaces can be challenging, especially for people who are not familiar with the layout or visiting for the first time. Hence, acquiring knowledge about the environment becomes necessary to successfully reach their preferred destination. In daily life, to acquire this knowledge, people either ask someone for directions or search for the preferred route on a map or the web. 
Meilinger \textit{et al.}~\cite{Meilinger2008} found that maps and verbal route descriptions are equally effective in conveying wayfinding knowledge, as they are memorized in a language-based format used mainly for wayfinding purposes. Hence, it is reasonable to assume that individuals can rely solely on verbal wayfinding instructions to achieve an effective navigation experience. But this experience would also depend on the type of route descriptions given. Inefficient or unclear route descriptions can lead to frustrations and errors in navigation.  

To address the challenge, the use of escort robots to guide people from their current positions to preferred destinations has become increasingly common in indoor environments (e.g., airports) \footnote[4]{https://airportimprovement.com/airport-robots-are-already-reality}. However, adoption of escort robots is limited primarily due to the following limitations -- (i)~cost of the robot due to their sophisticated design, mobility features, and escort capacity, (ii)~lack of scalability as one robot can escort only one person or group at a time resulting in inefficiency in crowded environments, (iii)~space requirements of the robot, and (iv)~limited adaptability to environmental factors like moving obstacles, dynamic layouts, and temporary barriers that are common in indoor spaces. Thus, stationary socially assistive robots that provide verbal route descriptions have emerged as an alternative due to their cost effectiveness, ability to assist multiple individuals in quick successions \& lower maintenance requirements. They also would face minimal disruptions caused by mechanical failures, thus ensuring reliable and efficient navigation support in diverse environments. 

A key element for a robot to effectively fulfill its role in providing route descriptions is the quality of the route descriptions provided. Many existing studies focus on natural language interaction with robots. They aim to build models for robots to understand human input~\cite{pramanick2019your, pramanick2022talk, pramanick2022doro} and how robots should navigate in public places~\cite{barua2020let, mavrogiannis2022social}. However, how a robot provides information to humans concerning route description is understudied. The question of what constitutes an ideal route description is not new~\cite{Daniel1998, Daniel2003}. Since people have cognitive limits in remembering and recalling verbal route descriptions, a long and detailed verbal description of the route may not be the best option. Hence determining the optimal length of verbal route descriptions that individuals can effectively remember and recall during indoor navigation is essential. 


This study aims to identify cognitive limitations and preferences in comprehending and retaining verbal route descriptions, establishing a measurable threshold for the ideal length of route descriptions through experiments. This study addresses two primary questions -- (1) How to define the concept of ``length'' in the context of route descriptions? (2) What is the optimal length of route descriptions that results in the best navigation performance when followed?

To address these questions, \textbf{firstly}, a theoretical framework is proposed by integrating findings from cognitive psychology and geospatial research to define a route segment corresponding to a chunk in working memory. By utilizing the concept of chunks in working memory~\cite{cowan2001}, this article proposes a comparison between the concept of a route segment and a chunk, aiming to define the `length' of a route description. Based on this comparison and the typical short-term capacity limit from cognitive psychology~\cite{cowan2001}, this paper hypothesizes that a similar limit of approximately four route segments would represent the optimal length of route description recallable during navigation. 

\textbf{Secondly}, based on this theoretical framework, an experiment is designed and conducted to assess people's navigational performance corresponding to route descriptions of varying lengths. Route descriptions of three different lengths of short, medium, and long are created for the same route, comprising 4, 6, and 8 route segments (or chunks in working memory), respectively. A correlation between superior navigation performance and the length of route descriptions confirms the theory that a route segment in geospatial research corresponds to a chunk in working memory. 

\textbf{Thirdly}, the methodology proposed in this work provides a template for future research investigating similar questions regarding route descriptions and navigational performance. 

\textbf{Finally}, This study identified promising trends, indicating that shorter route descriptions generally lead to improved navigational outcomes, aligning with the initial expectations.
\section{Theoretical framework}
\label{sec:related}
In this section, we establish a theoretical framework for this study. Baddeley \textit{et al.}~\cite{BADDELEY197447} shows working memory represents the ability to hold transient representations while simultaneously processing and assimilating ongoing events. A study by A M Hund~\cite{HUND2016233} found that visuospatial working memory plays an important role in wayfinding in an indoor environment. Hund's findings suggest that efforts aimed at facilitating wayfinding should consider the demands on working memory. This involves reducing the amount of information people have to remember as they navigate through the environment. The study emphasizes that the capacity limits of verbal working memory directly impact the process of recalling and following route descriptions in indoor settings. Individuals must temporarily hold and process verbal information in their working memory to successfully execute navigation tasks. Exceeding the capacity limits of verbal working memory may lead to difficulties recalling crucial landmarks, turns, or instructions, affecting the accuracy and efficiency of indoor navigation. Therefore, understanding and accommodating these limitations are essential for enhancing wayfinding instructions and improving user experience in indoor environments.

An early study by Miller~\cite{miller} suggested that people could remember about seven chunks in Short Term Memory (STM). Subsequent research indicated a more precise capacity limit of three to five chunks. Cowan in his seminal work on working memory capacity limits~\cite{cowan2001}, brought together a wide variety of data to suggest that the smaller capacity limit of four chunks is ideal. He defines a ``chunk'' as a collection of concepts with strong associations to one another and weaker associations to other concurrently used chunks. This sheds light on the cognitive processes underlying working memory capacity.

In geospatial research, most of the route description constitutes a list of triplets $<orientation, action, landmark>$~\cite{Denis1997}. This triplet forms the basis of a route description, which involves designating a landmark, reorienting the listener, and prescribing an action to continue the progression along the route. In order to understand the concept of chunks within the context of verbal route descriptions, it is important to understand how chunks are formed in memory. McLean and Gregg~\cite{mclean&gregg} described three ways chunks can be formed in working memory.
\begin{itemize}[leftmargin=*]
    \item  Some stimuli may already form a unit with which the subject is familiar. This means that individuals may naturally perceive certain elements as a cohesive unit because of their prior knowledge or familiarity with those elements. In the context of route descriptions, each segment of the route might be perceived as a chunk because people associate navigating each triplet $<orientation, action, landmark>$ as a sub-task.

    \item External punctuation of the stimuli may serve to create groupings of the individual elements. This refers to the use of external cues, markers, or punctuation to create groupings or chunks from individual elements. In the context of route descriptions, phrases like `first walk down', `then', `continue', etc. at the start/end of a route segment reinforce chunking. Additionally, pauses between route segments during the narration would further emphasize chunking. 

    \item The person may monitor his own performance and impose structure by selective attention, rehearsal, or other means.
    
\end{itemize}

Based on these prior works, we can conclude that the concept of a route segment fits closely with the idea of a chunk in working memory. Thus, this work suggests defining a chunk in the context of verbal route descriptions as a route segment corresponding to the triplet, $<orientation, action, landmark>$, as explained earlier. This proposed definition of chunk addresses the question of how to define the length of a route description.
Further, based on this definition of chunks as route segments and findings from cognitive psychology, we hypothesize that the ideal length of a route description is four route segments. This ideal length would facilitate easy recall and successful navigation completion. Together, this definition of chunk and the proposed optimal length collectively establish the theoretical framework for this study.

%% file: 02_methodology.tex
\section{Method}
\label{sec: method}
We conducted a user participation study where users navigated within a real-world environment. First, at the starting point of a given route, a participant watched videos of an AI-generated talking avatar narrating the directions to a destination. Then, the participant attempted to recall and follow those directions to reach the final destination. 

\subsection{Protocol}
The methodology used in this study was inspired by the research conducted by  Daniel \textit{et al.}~\cite{Daniel2003}, where similar approaches were utilized to examine the effectiveness of three types of route descriptions – labeled as good, poor, and skeletal – in guiding individuals to a specific location on campus. In the second phase of their experiment, participants were escorted to the starting position of the route, which did not overlap with the assigned route for the experiment. They were then randomly assigned one of the three route descriptions and instructed to study the printed route descriptions before initiating navigation. Participants were directed not to seek external assistance during the task, while the experimenter monitored from a distance of 5 meters, recording instances of three behavioral indicators: self-corrected errors, experimenter-corrected errors, and stops. The duration and locations of these indicators, along with the total navigation time, were also documented. 

We incorporate several elements from this procedure into our study, including conducting a navigation task to assess route description efficacy, leading participants to the starting position via a non-overlapping route, random assignment of route descriptions, instruction to participants not to seek help, recording the duration and locations of stops, as well as the total navigation time. 

However, our study differed significantly in several respects. In our study, participants were presented with pre-recorded videos containing the route descriptions, allowing them to listen to the instructions only once. This approach aimed to simulate real-world scenarios where users typically receive navigation instructions just once, such as when asking a robot for wayfinding guidance. Moreover, participants were provided with the option to terminate the experiment if they believed they could no longer recall the route description. To maintain our focus on navigation using the recall of route descriptions and considering the short duration of the selected routes, a 5-minute threshold was used. If participants exceeded this limit, the experimenter stopped the experiment as participants were likely wandering aimlessly and unable to reach the destination. Participants were also instructed to refrain from engaging with anyone, including the experimenter, during navigation, ensuring minimal distractions and a consistent experience for all participants. Additionally, the definition of errors as behavioral indicators was refined encompassing only errors, as experimenter-corrected errors were deemed inconsequential for our objective of evaluating navigation performance based on participants' route description recall. 

\subsection{Participants}
\begin{figure*}

\centering
\begin{subfigure}{0.24\textwidth}
    \centering
    \includegraphics[width=\textwidth]{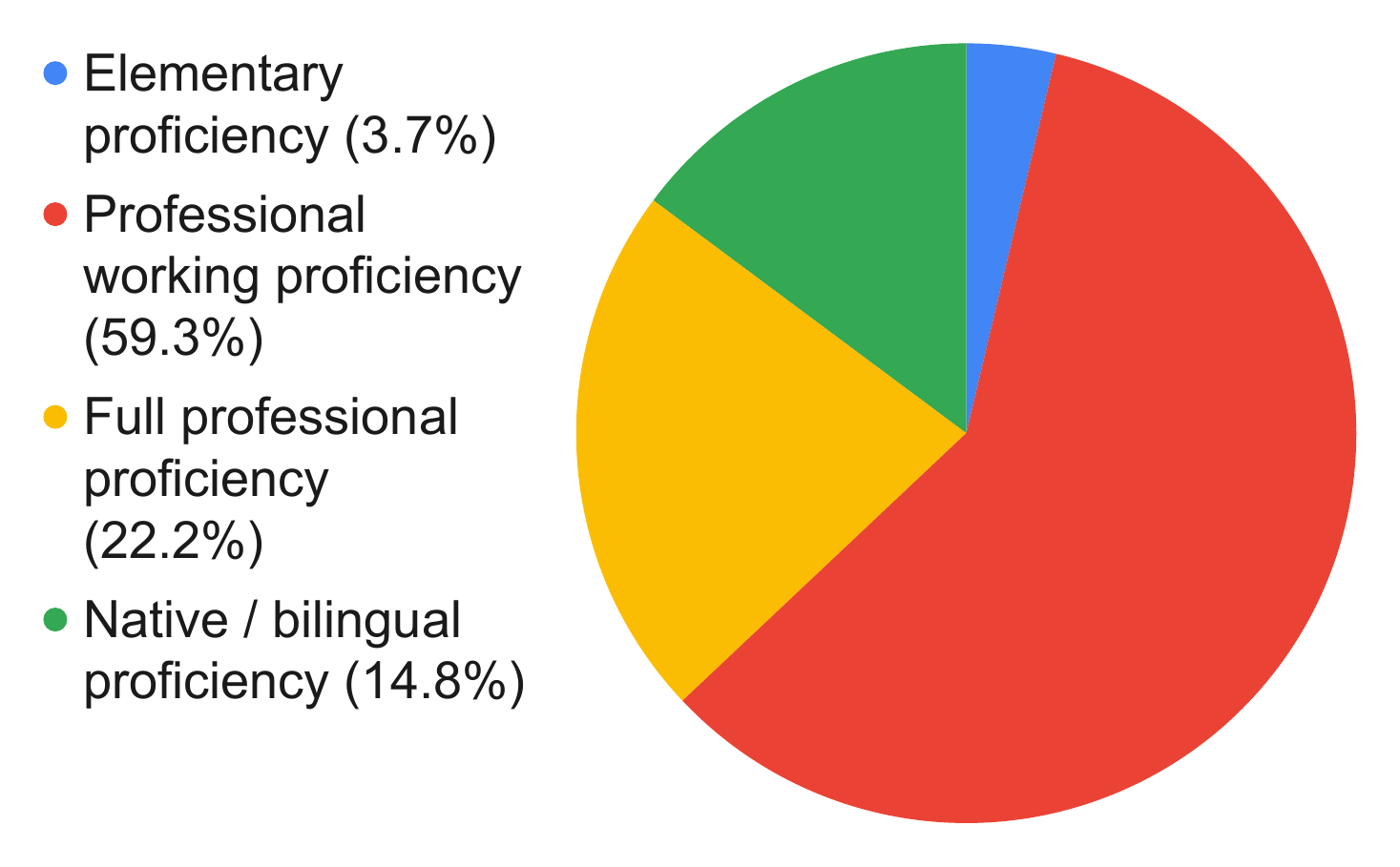}
    \caption{English \\proficiency }
    \label{fig:english-prof}
\end{subfigure}
\hfill
\begin{subfigure}{0.24\textwidth}
    \centering
    \includegraphics[width=\textwidth]{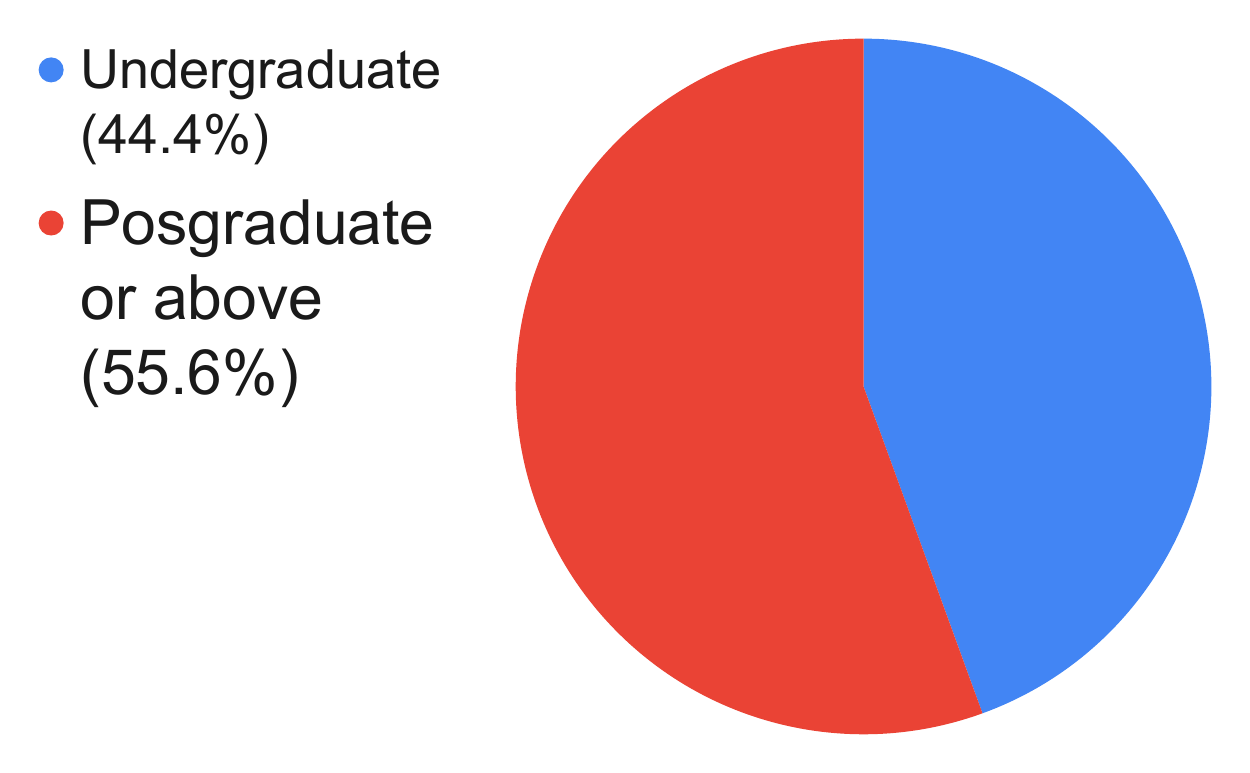}
    \caption{educational \\qualification }
    \label{fig:edu-qual}
\end{subfigure}
\hfill
\begin{subfigure}{0.24\textwidth}
    \centering
    \includegraphics[width=\textwidth]{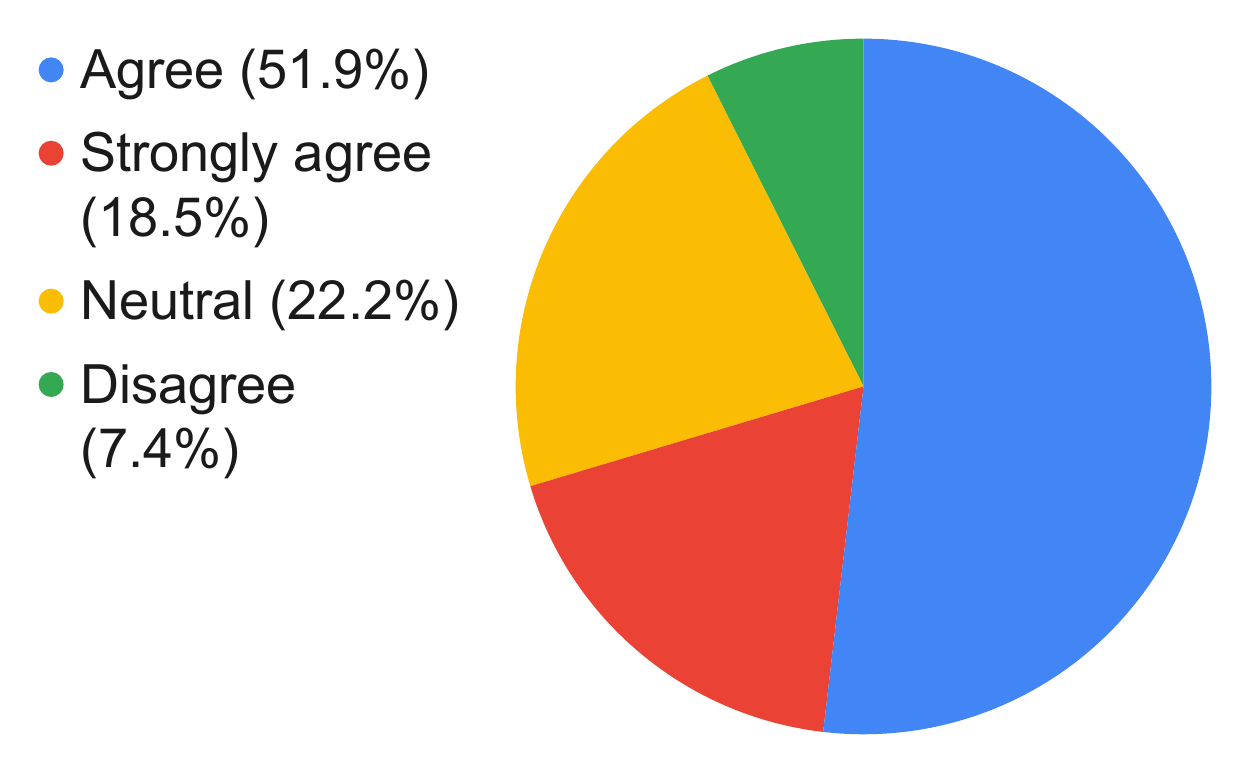}
    \caption{comfortable with voice based navigation assistance}
    \label{fig:nav-assitants}
\end{subfigure}        
\hfill
\begin{subfigure}{0.24\textwidth}
    \centering
    \includegraphics[width=\textwidth]{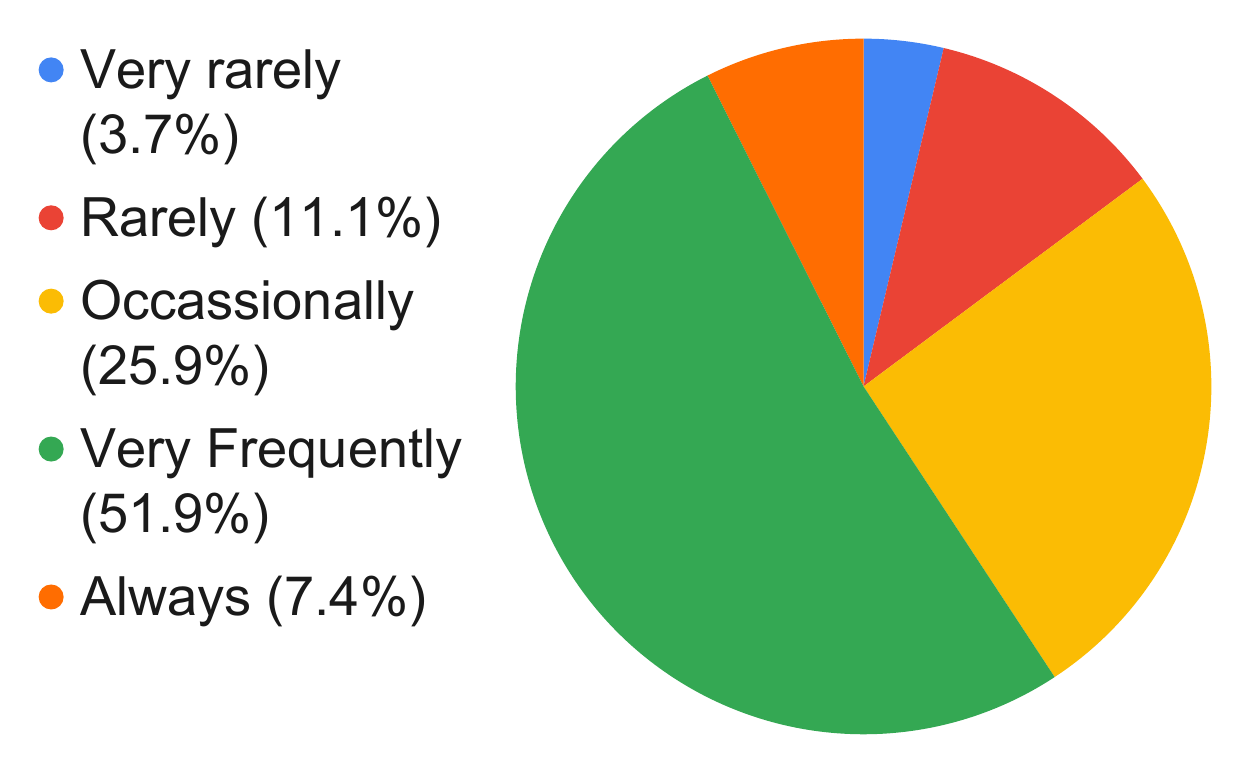}
    \caption{frequency of using \\navigation app}
    \label{fig:freq-nav-apps}
\end{subfigure}
\caption{Statistics of participants' background on various aspects.}
\label{fig:participant_background}
\end{figure*}

A total of 27 participants were part of this study, who are colleagues of the author, but do not belong to the same research group or research domain. The participants included 20 males (mean age = 26.10, SD = 4.27) and 7 females (mean age = 23.43, SD = 4.15). The participants were all non-native English speakers and had different levels of English proficiency according to their own ratings (Fig.~\ref{fig:english-prof}). All the participants had at least an undergraduate degree, out of which 55\% had a postgraduate degree or above (Fig.~\ref{fig:edu-qual}). All the participants reported that they are familiar with at least one smartphone map application, while a few reported familiarity with some other navigation apps too. Most participants (51.9\%) also agreed that they are comfortable with voice-based navigation assistants (Fig.~\ref{fig:nav-assitants}). Also, most (51.9\%) of the participants were frequent users of navigation apps or took the help of verbal instructions during navigation (Fig.~\ref{fig:freq-nav-apps}). All the participants were visiting the building where the experiment took place for the first time and none of them had any previous knowledge of the environment. All the participants signed a consent form before participating in the experiment and were presented with a small token of appreciation as an acknowledgment of their participation.

\subsection{Experiment site}
The environment used for this study was an office building. This particular building was selected due to its unfamiliarity to the participants prior to the experiment. Four routes were carefully selected, all of comparable lengths and complexities, ensuring participants could be guided to their starting points without route overlap. As the current study aimed to assess the impact of different lengths of route descriptions on navigational performance for the same route, it was necessary to ensure comparability across the four routes in several aspects. None of the routes traversed any floor or area more than once, each covering entirely distinct, non-overlapping areas. Moreover, all routes involved a change in elevation, either through the use of stairs or elevators.
The four routes chosen are described in (Table~\ref{tab:table2}). The time it took for the experimenter to walk each route from the starting point to the destination at a normal pace ranged from 1 minute 15 seconds to 1 minute 30 seconds, without accounting for the wait time for the lift, if any, indicating a uniformity in the lengths of these four routes. 

\begin{table}[]
\centering
\begin{tabular} { 
    | m{0.9cm} | m{6.3cm}|}
        \hline
        Route name & Definition  \\
        \hline
        {Route 1 (R1)}  & Begins at the entrance to the building, goes up a flight of
stairs, and end in front of a designated room on the first
floor \\
        \hline
        Route 2 (R2)
        & Begins in front of a training room on the
second floor, goes up six floors via the lift, and ends at the
table tennis court on the eighth floor. \\    
        \hline
        Route 3 (R3)
        &  Begins in
front of an office space on the seventh floor, goes down four
floors via the lift, and ends in front of a server room on the
third floor.\\ 
\hline
Route 4 (R4) & Begins in front of an office space on the fifth floor, goes up one flight of stairs, and ends in front of a designated office space on the sixth floor. \\
    \hline
    \end{tabular}  
\caption{Four routes chosen for the experiment.}
  \label{tab:table2}
\end{table}

\subsection{Creation of route descriptions}
For each of the four routes, three route descriptions of different lengths in terms of the number of route segments -- short, medium, and long, corresponding to 4, 6, and 8 route segments were created. 

First, a list of all route segments describing the entire route was generated. This was done in accordance with guidelines proposed by M. Denis~\cite{Denis1997}, which emphasize the importance of -- (a)~limiting the number of statements, (b)~avoiding redundancy and over-specification, (c)~referencing visible, permanent, and relevant landmarks, (d)~prioritizing determinate descriptions, and (e)~clearly articulating associations between actions and landmarks. 

Then, route descriptions of three different lengths were generated for the same route using seven rules of chunking derived from Klippel's spatial chunking rules for outdoor environments ~\cite{Klippel2009}, but modified to fit into the context of a complex indoor environment.
The goal of applying chunking rules was to cluster some parts of a given route so as to shorten it without losing important information needed to reach the destination. The definitions of these rules in the context of an indoor environment are listed below.
\begin{enumerate}[leftmargin=*]
    \item \textit{Numerical chunking }: counts decision points like corridor intersections or turns to summarize them into a single instruction, e.g., ``take the second turn''. 

    \item \textit{Chunking based on structural features}: involves identifying and utilizing salient structural characteristics to describe a route, e.g., ``walk till the escalators''.

    \item \textit{Local chunking based on point-like landmarks}: uses distinctive objects or points as a reference, e.g., ``walk past the restroom''.

    \item \textit{Local chunking based on linear landmarks}: uses linear features like hallways and designated pathways to guide the navigation along a path, e.g., ``follow the corridor''.

    \item \textit{Local chunking based on area-like landmarks along the route}: involves using landmarks with area-like features that spread along the route and are relevant for multiple decision points, like halls, atrium, lobbies, etc., e.g., ``pass through the hallway''.

    \item \textit{Global chunking based on point-like landmarks}: involves providing route descriptions that do not fully specify the entire route but rely on well-known or easily identifiable landmarks to guide the Wayfinder, e.g., ``go to the second floor''.

    \item \textit{Global chunking based on area-like landmarks}: uses area-like landmarks to guide navigation and potentially allow for chunking large parts of the route without the need to announce intermediate decision points, e.g., ``get to the other side of the atrium''.
\end{enumerate}

Initially, the lists comprised 14, 11, 11, and 10 route segments for R1, R2, R3, and R4, respectively. Subsequently, the chunking principles were employed to select 4, 6, and 8 route segments corresponding to short, medium, and long route descriptions, respectively. This process yielded a total of 12 route descriptions. A video was created for each of the 12 route descriptions, featuring an AI talking head~\footnote[5]{https://www.colossyan.com/} delivering narration in a general American accent, commonly recognized by most people. A 2-second pause separates each route segment, allowing listeners to perceive them as distinct chunks. These videos were stored in a tablet that the participants used during the experiment. Each participant engaged in three navigation tasks (trials) randomly selected from the 12 route descriptions, such that each of them got one short, one medium, and one long route description corresponding to random routes in random order. As mentioned earlier, each task began by watching the corresponding pre-recorded video.


\subsection{Questionnaire}
We designed a structured questionnaire to gather subjective feedback from participants regarding each trial and their overall experience comprising five sections.
\begin{itemize}
    \item The initial section provides generic instructions to the participant, a consent form, and demographic questions covering age, gender, navigation app usage frequency, comfort with navigation assistants, self-rated English fluency, etc.
    
    \item The subsequent three sections were completed after each trial, focusing on the participants' subjective analysis of the route description, including the perceived difficulty level of the navigation and confidence levels experienced during navigation.

    \item The fifth section provided an overall feedback form, enabling participants to compare the three trials, assess the perceived difficulty in following the route descriptions, and compare the lengths of routes.
    
\end{itemize}

\subsection{Procedure} 
Each participant completed three trials involving short, medium, and long route descriptions for various routes, with the order randomized. Routes for each trial were also randomly assigned. Each trial involved listening to the recorded descriptions, navigating from start to finish, and providing feedback on the navigation experience. This resulted in three experiment groups, each with 27 participants following different route lengths.

The experiment took place in a building next to the workplace of the participants. The participants were guided there by the experimenter.  At the entrance of the building, the participant was given a note to brief them about the experiment and what to expect. After reading and signing the consent form, the experimenter then took the participant to the starting point of the route corresponding to the first trial. Here the participant was positioned at the starting point in the pre-defined orientation, then informed of the name and corresponding floor number of their starting position (their current position) and the name of their destination. They were then given the route description video, which they watched only once. After listening to the route description, the participant said `start' and started navigating along the route they had memorized. The participants were warned not to ask for help either from the experimenter or from other pedestrians. The participants were informed they could say stop to halt the navigation in two cases -- if they think they have reached the destination, and if they think they are no longer able to recall the route and are simply going around the building searching for the destination. They were informed that in case the navigation took more than 5 mins, the experimenter would say stop and halt the navigation.

During navigation, the experimenter followed the participant at a distance of 5 m and noted any occurrences of the two behavioral indicators: stops and errors. A stop was defined as a pause during walking, and an error was defined as a deviation from the original route followed by a return to the route.  The duration and location of these indicators were noted, alongside the total navigation duration and success/failure status. After each trial, participants rated the difficulty of following spoken route descriptions on a scale of 1 to 10 and their confidence levels on a scale of 1 to 5. Then the participant was taken to the starting point of the next trial through a non-overlapping route.
At the end of all three trials, the participant also provided their subjective feedback on some questions comparing the three route descriptions and their relative difficulty and perceived length. 

\subsection{Hypothesis}
The main hypotheses of this study are listed below. 
\begin{itemize}[leftmargin=*]

    \item \textbf{H1}: Participants navigating using short route descriptions will demonstrate higher successful completion rates compared to those navigating using medium or long route descriptions.
    \item \textbf{H2}: Participants navigating using short route descriptions will demonstrate shorter total navigation durations compared to those navigating using medium or long route descriptions. 
    \item \textbf{H3}: Participants navigating using short route descriptions will exhibit shorter durations of errors and stops compared to those navigating using medium or long route descriptions.
    \item \textbf{H4}: Participants navigating using short route descriptions will make a smaller number of errors and will stop fewer times compared to those navigating using medium or long route descriptions.
    
\end{itemize}

%% file: 03_results.tex
\section{Results and Discussion}
As mentioned earlier, navigation tests were conducted for 27 participants, each completing 3 trials (one short, medium, and long route descriptions). Data for one medium-length trial was removed due to an interruption in the experiment, resulting in 80 total trials. Out of these, 58 trials resulted in successful navigation, while 22 were unsuccessful due to various reasons such as incorrect route selection or disorientation of the participants. Data from unsuccessful trials were excluded from time measurements and behavioral indicators analysis.

\begin{figure*}

\centering
\begin{subfigure}{0.24\textwidth}
    \centering
    \includegraphics[width=\textwidth]{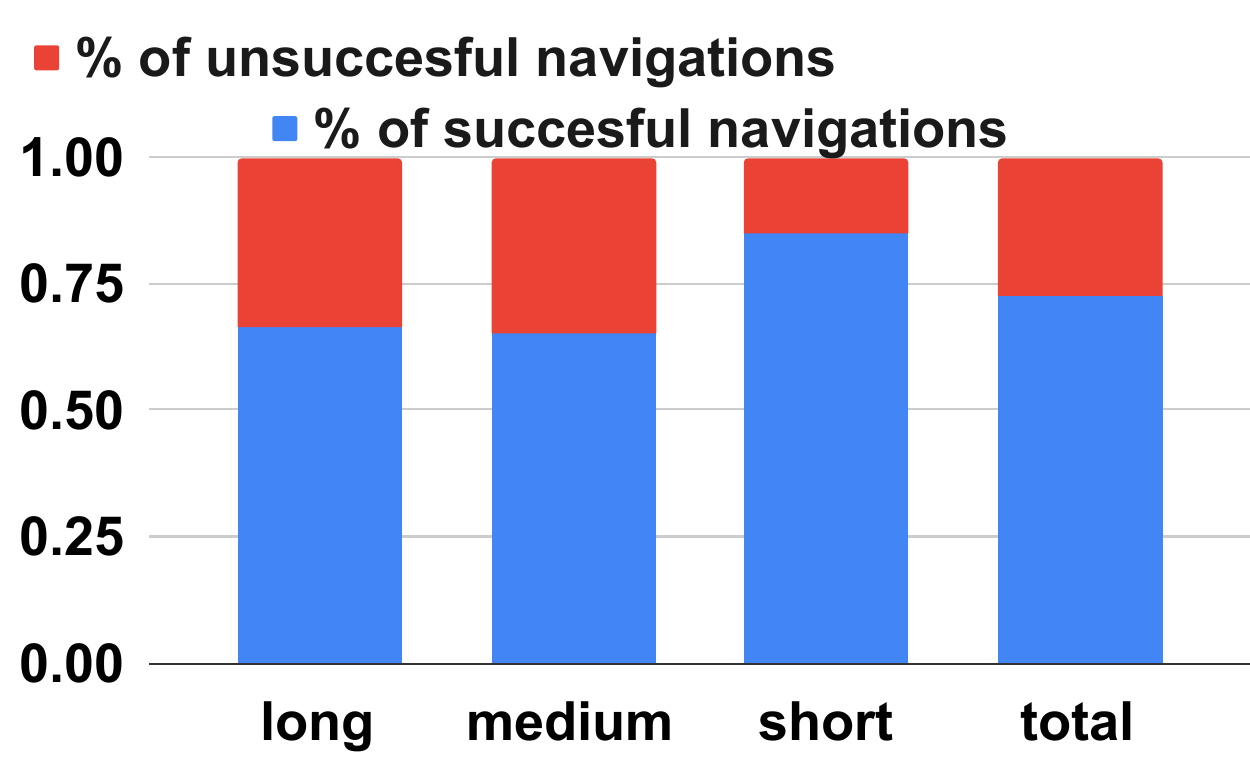}
    \caption{navigation \\success rates}
    \label{fig:succes-rate-plot}
\end{subfigure}
\hfill
\begin{subfigure}{0.24\textwidth}
    \centering
    \includegraphics[width=\textwidth]{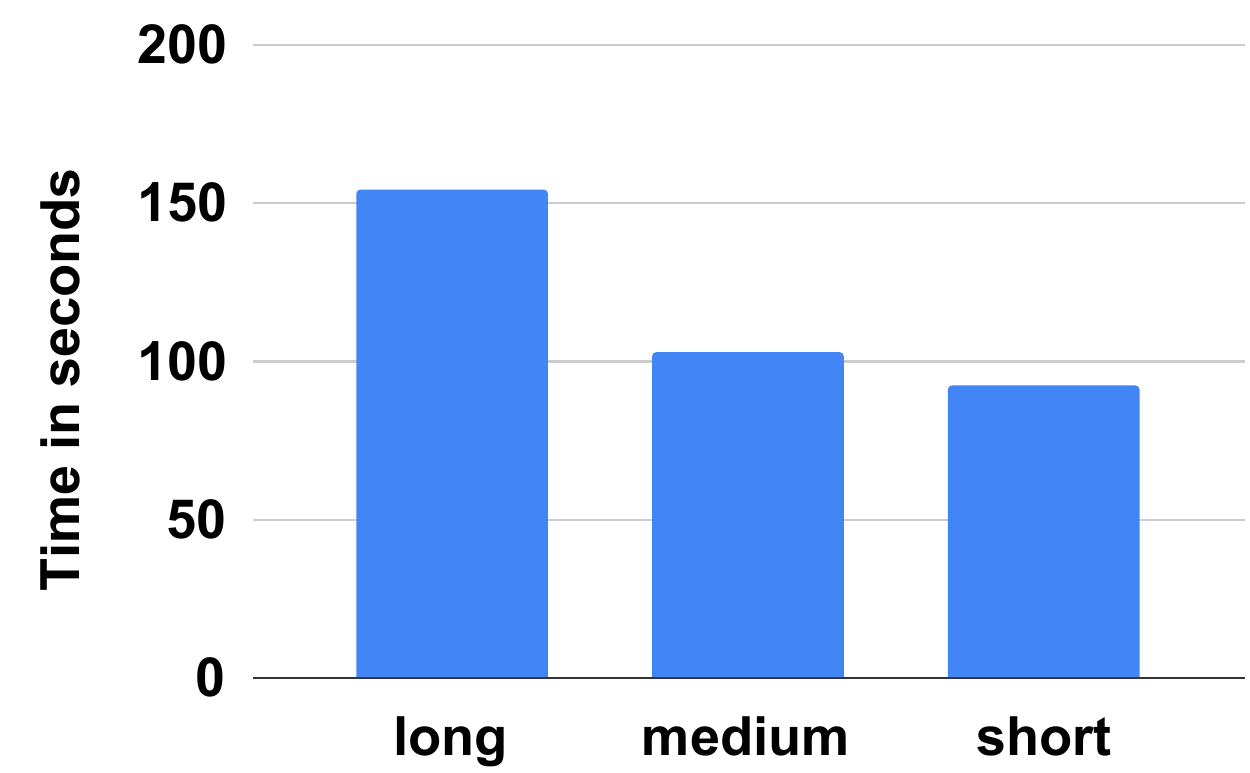}
        \caption{mean total \\duration of navigation}
    \label{fig:Mean-total-duration}
\end{subfigure}
\hfill
\begin{subfigure}{0.24\textwidth}
    \centering
    \includegraphics[width=\textwidth]{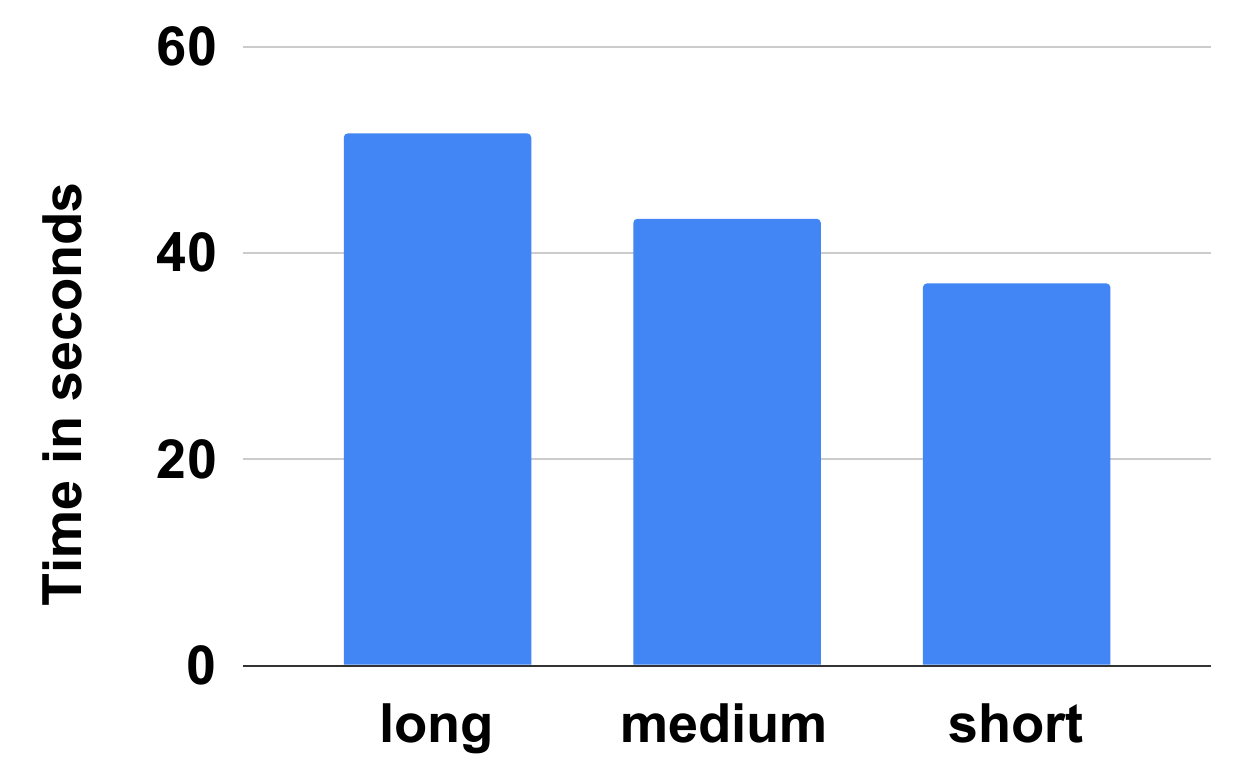}
    \caption{mean total \\duration of errors}
    \label{fig:duration-of-errors}
\end{subfigure}
\hfill
\begin{subfigure}{0.24\textwidth}
    \centering
    \includegraphics[width=\textwidth]{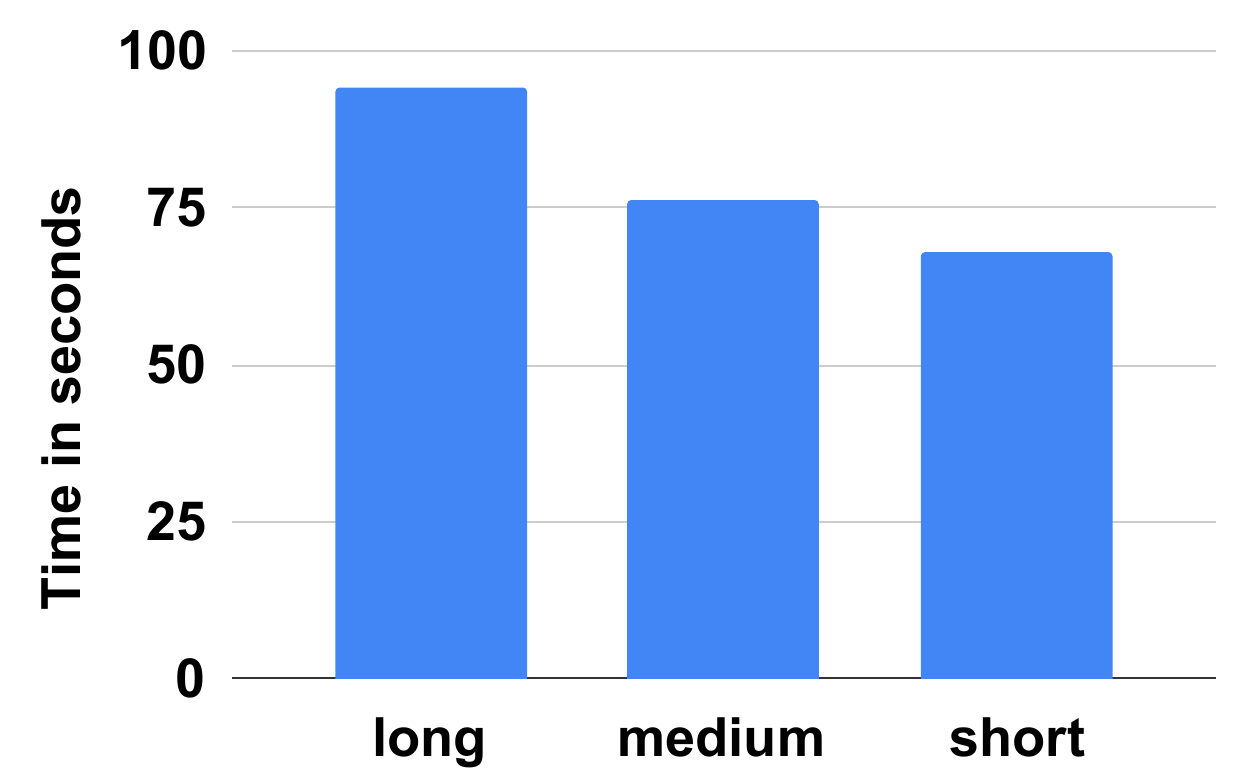}
    \caption{mean actual \\navigation times}
    \label{fig:actual-nav-time}
\end{subfigure}        
\caption{Plots of time measurements and behavioral indicators across different route description lengths.}
\label{fig:result-plots-1}
\end{figure*}

\subsection{Navigation success rate}
We ran a chi-square test to explore the potential relationship between the length of route descriptions and the success rate of navigation. Despite the lack of statistical significance revealed by the test  
$ \chi^2 (2)=3.30, p=.192 $, a closer examination reveals interesting patterns. From Fig.~\ref{fig:succes-rate-plot}, it is apparent that short route descriptions exhibit a greater frequency of successful navigations relative to other route descriptions coupled with a lower frequency of unsuccessful navigations. Specifically, short, medium, and long route descriptions resulted in success rates of 85\%, 65\%, and 67\%, respectively. Remarkably, the success rate of 85\% for short route descriptions surpassed the overall success rate of 72.5\%. It is plausible that the ease of recall of short route descriptions contributes to the higher success rates of navigation, aligning with our hypothesis (H1). 


\subsection{Time measurements}
\label{sec:time-measurements}
Here, the results related to the total duration of navigation, duration of errors, duration of stops, and actual navigation times are listed.

\subsubsection{Total duration of navigation}

A one-way ANOVA~\cite{girden1992anova} performed to compare the effect of three different route descriptions on total duration of navigation revealed that there was a statistically significant difference in mean total duration between at least two groups (F(2, 59)=[7.15], p=.002). Tukey’s HSD Test~\cite{tukeyhsd} for multiple comparisons found that the mean value of total duration was significantly different between the long \& medium route descriptions (p=.015, 95\% C.I.=[8.52, 93.27]), and the long \& short route descriptions (p=.002, 95\% C.I.=[19.90, 103.62]). There was no statistical significance for the mean total duration of navigation between medium and short route descriptions (p=.812). 

According to an independent sample t-test, participants took longer time to complete the navigation for long route descriptions (M=154.10, SD=83.69) than for medium route descriptions (M=103.20, SD=28.10), t(39)=2.58, p=.014. The participants also took longer time for long route descriptions (M=154.10, SD=83.69) than for short route descriptions (M=92.33, SD=40.44), t(40)=3.05, p=.004 (Fig.~\ref{fig:Mean-total-duration}). This trend aligns with our hypothesis (H2) as short route descriptions correspond to the shortest durations of navigation. To understand why participants who followed the long route descriptions took longer than others, the time measurements of the duration of stops and duration of errors are examined. 

\subsubsection{Duration of errors}
The data failed the Shapiro-Wilk test~\cite{shapiro-wilk1,shapiro-wilk2} of normality ($p <.001$), hence the non-parametric Kruskal-Wallis Test~\cite{kruskal-wallis} was conducted to examine the differences on the duration of errors for different lengths of route descriptions. No significant differences (Chi square=1.27, p=.53, df=2) were found among the three categories of route description lengths (short, medium, and long). Interestingly, a comparison of the mean duration of errors for the three route description lengths [Figure ~\ref{fig:duration-of-errors}] shows that the mean duration of errors increased as the route description length increased. Specifically, the mean duration of errors for short route descriptions (M=37.03, SD=45.8) was smaller than those for medium (M=43.23, SD=64.24), which was smaller than long (M= 51.59, SD=52.96) route descriptions. These preliminary findings suggest the possibility that short route descriptions are associated with shorter durations of errors, potentially making them more conducive to efficient navigation. These emerging trends align with our expectations (H3) and warrant further exploration in future studies.


\subsubsection{Duration of stops}
The data failed the Shapiro-Wilk test of normality ($p <.001$), hence the non-parametric Kruskal-Wallis Test was conducted to examine the differences in duration of stops for different lengths of route descriptions. No significant differences (Chi square=.34, p=.845, df =2 ) were found among the three categories of route description lengths (short, medium, and long). This was expected, as stopping behavior during navigation is subjective and may vary among individuals, potentially explaining the lack of significance in the results.

\subsubsection{Actual navigation time}
While the findings related to the total duration of navigations and duration of errors align with our expectations, it is possible that the variations in individuals' walking speeds while navigating according to route descriptions of varying lengths could impact this metric. To address this concern, potential differences in walking pace and their impact on navigational performance were examined. Stop and error durations for each participant were calculated and these were subtracted from total navigation duration. An ANOVA showed a significant effect of route description length on actual navigation times (F(2,58)=5.16, p=.009), suggesting that differences in navigation durations could be attributed not only to the duration of errors but also to variations in walking paces. A comparison of mean actual navigation times shows that participants tended to take less time to navigate using short route descriptions (M=67.83, SD=24.45) than using medium (M=76.35, SD=23.67), or long route descriptions (M=94.28, SD=30.86). 

Additionally, there was a need to verify that these differences in time measurements were not associated with variations between routes. The goal was to assess the impact of different routes (R1, R2, R3, R4) on total duration, error duration, and stop duration across all three route description lengths. For this, the durations were separately analyzed for each route length category. A Kruskal-Wallis test performed to compare the effects of only short route descriptions corresponding to four different routes (R1, R2, R3, R4) on the total duration of navigation, in Table~\ref{tab:table1}, revealed that there was a significant difference between the four routes. Note that no short descriptions for R3 were analyzed due to failed navigation attempts. This would have influenced the differences in total duration among the routes. In contrast, the Kruskal-Wallis tests (Table~\ref{tab:table1}) showed no significant effect of different routes on total duration, error duration, or stop duration for route descriptions of different lengths. Based on this analysis, this work establishes that differences in time measurement values were not due to differences between the four routes, suggesting their similarity.

\begin{table}[]

\centering
\begin{tabular} { 
    | m{1.2cm} | m{2cm} | m{1.8cm} | m{1.2cm} |}
        \hline
        Route description length & Parameter & Kruskal-Wallis test results & Significance result  \\
        \hline
        \hline
        \multirow{3}{1.2cm}{Short (four chunks)}  & total duration of navigation & $\chi^2$=6.54, p=.038, df=2 & Significant \\
        \cline{2-4}
        & duration of errors & $\chi^2$=2.99, p=.225, df=2 & Not significant \\ 
         \cline{2-4}
         & duration of stops & $\chi^2$=2.79 , p=.247, df=2 & Not significant \\
        \hline
        \multirow{3}{0.8cm}{Medium (six chunks)}  & total duration of navigation & $\chi^2$=3.17, p=.366, df=3 & Not significant  \\
         \cline{2-4}
         & duration of errors & $\chi^2$=3.68, p=.298, df=3 & Not significant \\ 
         \cline{2-4}
         & duration of stops & $\chi^2$=2.35, p=.504, df=3 & Not significant \\
        \hline
        \multirow{3}{1.2cm}{Long (eight chunks)} & total duration of navigation & $\chi^2$=6.87, p=.076, df=3 & Not significant \\
         \cline{2-4}
         & duration of errors & $\chi^2$=4.85, p=.183, df=3 & Not significant \\ 
         \cline{2-4}
         & duration of stops & $\chi^2$=1.91 , p=.592, df=3 & Not significant \\
    \hline
    \end{tabular}  
\caption{Statistical tests for effect of four routes on different parameters.}
  \label{tab:table1}
\end{table}

Since the same participants were involved in all three experimental groups, the variations in actual navigation times cannot be attributed to individual differences in inherent walking speeds. However, based on the results discussed here, variations in actual navigation times can only be linked to pace differences between the three groups. Hence, it seems plausible that participants walked faster while using short route descriptions compared to long or medium route descriptions, despite navigating similar routes.


\subsection{Behavioral indicators}
\label{sec:freq-behavioral-indicators}
Here the analysis of self-reported confidence ratings, perceived difficulty ratings, number of errors, and number of stops are listed.

\subsubsection{Self reported confidence ratings}
Participants rated their confidence levels while navigating with given route descriptions on a scale of 1 to 5. This data failed the Shapiro-Wilk test of normality ($p <.001$). Kruskal-Wallis test to examine the differences in self-reported confidence ratings for different lengths of route descriptions showed no significant differences (Chi square=.95, p=.623, df=2). Interestingly, mean confidence ratings decreased with route description length, suggesting higher confidence while following short route descriptions (M=4.09, SD=0.73) compared to medium (M=3.88, SD=0.99) and long route descriptions (M=3.78, SD=0.94). 

\subsubsection{Perceived difficulty ratings}
These were on a scale of 1 to 10, where 1 for the least difficulty and 10 for the most difficulty while navigating using the given route description. This data failed the Shapiro-Wilk test of normality ($p <.001$). Kruskal-Wallis test to examine the differences in perceived difficulty ratings for different lengths of route descriptions showed no significant differences (Chi square=.95, p=.623, df=2). 
However, mean difficulty ratings increased with the length of route descriptions. The mean difficulty rating for short route descriptions was smaller (M=4.52, SD=0.64) than those for medium (M=4.64, SD=0.63) and long (M=5.39, SD=0.69). 

\subsubsection{Number of errors}
The data failed the Shapiro-Wilk test of normality ($p <.001$), hence the non-parametric Kruskal-Wallis Test was conducted to examine the differences in the number of errors for different lengths of route descriptions. No significant differences (Chi square=5.65, p=.059, df=2) were found among the three categories of route description lengths (short, medium, and long). Also, out of 58 data points, 44 were zero values, indicating that most participants made no errors, which accounts for the test results.

\subsubsection{Number of stops}
The data failed the Shapiro-Wilk test of normality ($p <.001$), hence the non-parametric Kruskal-Wallis Test was conducted to examine the differences in the number of stops for different lengths of route descriptions. No significant differences (Chi square=.58, p=.748, df=2) were found among the three categories of route description lengths (short, medium, and long). 
This outcome is not surprising given the subjective nature of stopping behavior during navigation. The lack of significance could be attributed to the inherent subjectivity of this trait.

\subsection{Discussion}
The results suggest that short route descriptions demonstrated a higher frequency of successful navigation compared to the other two (H1). The analysis also revealed significant differences in total navigation duration among the three route description lengths, with short route descriptions showing the shortest duration, as hypothesized (H2). The findings from the duration of errors analysis suggest that short route descriptions are associated with shorter duration of error (H3). 
The analysis of the number of stops and errors did not reveal any relationship with route description length, thus our hypothesis (H4) could not be supported.
Additionally, our analysis of actual navigation times showed that participants navigated shorter routes more quickly compared to longer or medium routes. Analysis of perceived difficulty and confidence level suggest that short route descriptions were perceived as least difficult to follow and induced the most confidence in participants. This aligns with our expectation of a more favorable navigation experience with short descriptions.


%% file: 05_future_work.tex
\section{Limitations and Future Work}
One primary limitation of our study was the relatively small sample size, potentially reducing statistical power. 
Additionally, the lack of a fully controlled environment for the experiment led to delays and external distractions, which could have increased the variability of data. Future studies could use more controlled settings like simulations to ensure more accurate results.
Our measurements were prone to experimenter biases and errors, given that a single experimenter was responsible for collecting all the data and all parameters were manually recorded by the experimenter. Future research could involve multiple experimenters or utilize automated data collection methods, such as sensors to accurately track distances walked, or cameras to record the navigation.
Another important direction for future studies would be to explore other experiment methods to complement the findings of the navigation experiment. One potential approach could be to incorporate a recall-only test to assess participants' ability to remember route descriptions under normal circumstances.
Future studies could also explore measuring more parameters to judge the effectiveness of the navigation. For example, the heart rate could be measured to analyze the stress level or cognitive load during navigation, and observable behaviors like changes in walking speed, hesitation, facial expression, etc. could be measured to understand at what points participants are finding it difficult to navigate. 
Addressing these limitations will be crucial for advancing our understanding of route descriptions and navigational performances, ultimately leading to more effective navigation systems and strategies.

%% file: 04_conclusion.tex
\section{Conclusion}
The purpose of this study was to provide empirical data regarding the effects of route description lengths on navigational performance. Route descriptions of different lengths (short, medium, long) were created for routes of similar lengths (R1, R2, R3, R4) comprising 4, 6, and 8 route segments, respectively. The results provided initial insights and the emerging trends aligned with our expectation that individuals following short route descriptions would exhibit superior navigational performances. This implies that the expected ideal length of route descriptions could be four segments, corresponding to the short route descriptions. Our study also found that shorter route descriptions could be related to faster walking paces, along with higher confidence and lower difficulty experienced by people while navigating. Overall, while further research is warranted to confirm our findings and explore underlying relations, the observed trends provide foundations for future investigations. Incorporating these insights into the design of navigation systems could potentially enhance user experiences.
